\documentclass[journal]{IEEEtran}

\usepackage[numbers,sort&compress]{natbib} 
\usepackage{float}

\usepackage{multirow}
\usepackage{threeparttable}

\usepackage[ruled,linesnumbered]{algorithm2e} 
\usepackage{amsmath}
\usepackage{amsfonts}
\usepackage{amssymb}  

\usepackage{graphicx}
\usepackage{stfloats}

\graphicspath{{images/}}                     

\usepackage[hidelinks]{hyperref}

\usepackage[utf8]{inputenc}

\usepackage{pifont}

\usepackage{booktabs}

\usepackage{soul}
\usepackage{color, xcolor}
\soulregister\cite7
\soulregister\citep7
\soulregister\citet7 
\soulregister\ref7 
\soulregister\pageref7 

\begin{document}
%


\title{A Late-Stage Bitemporal Feature Fusion Network for Semantic Change Detection}

\author{Chenyao Zhou, Haotian Zhang, Han Guo, \\Zhengxia Zou, \IEEEmembership{Member,~IEEE}, and Zhenwei Shi$^\star$,  
\IEEEmembership{Senior Member,~IEEE \vspace{-1em}}

\thanks{
The work was supported by the National Natural Science Foundation of China under the Grants 62125102, the National Key Research and Development Program of China (Grant No. 2022ZD0160401), the Beijing Natural Science Foundation under Grant JL23005, and the Fundamental Research Funds for the Central Universities.
\emph{(Corresponding author: Zhenwei Shi (e-mail: shizhenwei@buaa.edu.cn))}}

\thanks{Chenyao Zhou, Haotian Zhang, Han Guo, and Zhenwei Shi are with the Image Processing Center, School of Astronautics, Beihang University, Beijing 100191, China, and with the State Key Laboratory of Virtual Reality Technology and Systems, Beihang University, Beijing 100191, China, and also with the Shanghai Artificial Intelligence Laboratory, Shanghai 200232, China.
}

\thanks{Zhengxia Zou is with the Department of Guidance, Navigation and Control, School of Astronautics, Beihang University, Beijing 100191, China, and also with the Shanghai Artificial Intelligence Laboratory, Shanghai 200232, China.}
}
\date{May. 2024}

\maketitle
\begin{abstract}
Semantic change detection is an important task in geoscience and earth observation. By producing a semantic change map for each temporal phase, both the land use land cover categories and change information can be interpreted. Recently some multi-task learning based semantic change detection methods have been proposed to decompose the task into semantic segmentation and binary change detection subtasks. However, previous works comprise triple branches in an entangled manner, which may not be optimal and hard to adopt foundation models. Besides, lacking explicit refinement of bitemporal features during fusion may cause low accuracy. In this letter, we propose a novel late-stage bitemporal feature fusion network to address the issue. Specifically, we propose local global attentional aggregation module to strengthen feature fusion, and propose local global context enhancement module to highlight pivotal semantics. Comprehensive experiments are conducted on two public datasets, including SECOND and Landsat-SCD. Quantitative and qualitative results show that our proposed model achieves new state-of-the-art performance on both datasets.
\end{abstract}


\IEEEpeerreviewmaketitle


\section{Introduction}
\label{sec:introduction}

\IEEEPARstart{R}{emote} sensing imagery interpretation plays an important role in geoscience and earth observation. As a fundamental task, semantic segmentation (SS) aims to classify pixels in remote sensing images into distinct land use land cover (LULC) categories for surface mapping. To better understand urbanization and its impact on environmental evolution, binary change detection (BCD) have been developed to monitor the changed regions among different temporal phases by predicting a binary mask \cite{bruzzone2012novel, jiang2022survey}. To further elevate the coarse-grained change occurrence mapping into fine-grained ``from-to'' semantic transition correspondence \cite{BiSRNet}, semantic change detection (SCD) techniques are receiving increasing attention in recent literature. By generating a semantic mask for each temporal phase containing not only the change/no change information but also the detailed LULC semantics within this particular temporal phase, richer change context can be demonstrated.

\begin{figure}[tbp]
    \vspace{-0em}
        \centering
        \includegraphics[width=0.485\textwidth]{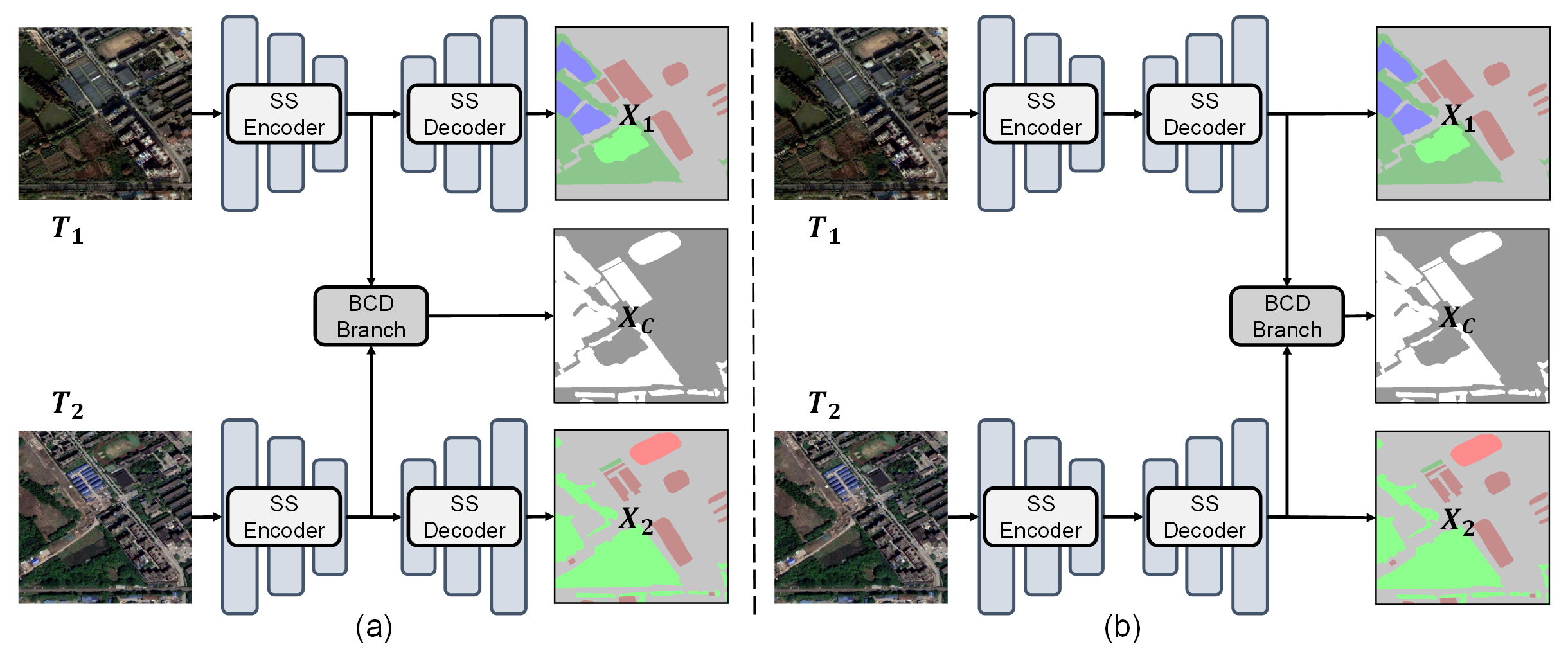}
        \caption{Architecture comparison between previous works and our proposed model. (a) Previous works merge bitemporal SS branches from encoders. (b) Our proposed network fuse SS decoded features to achieve BCD. 
        }
        \label{fig:architecture}
    \vspace{-1em}
\end{figure}

Before the prevalence of deep learning, traditional change detection methods adopt handcrafted features with the help of algebra, statistics and transformation \cite{parelius2023review}. With the intrinsic modeling ability of deep learning based algorithm, considerable improvements have been made mainly in the scope of bitemporal input SCD. By regarding SCD as a SS task for each temporal phase with additional ``no-change'' category, some CNN based and Transformer based siamese networks are implemented in an end-to-end manner \cite{daudt2018fully, SCDNet, Landsat}. However, without explicit constraint to regulate the change region mapping within each temporal branch, these methods struggle to suppress the changed regions discrepancy. To align together changed regions of different temporal phases, recently some multi-task learning based networks with triple branches are proposed to separately learning the LULC semantics within each temporal phase and the change location across time interval \cite{daudt2019multitaskHRSCD, BiSRNet, ding2024joint, DecoderFocus}. In this scenario, two SS branches are developed to model the LULC semantics for bitemporal inputs, whilst a BCD branch is specificly designed to capture the change context. The predicted binary change mask is then utilized to filter out all the unchanged regions in predicted bitemporal semantic maps through dot product, resulting in the final predicted bitemporal semantic change maps.

\begin{figure*}[htbp]
    \vspace{-1em}
        \centering
        \includegraphics[width=0.9\textwidth]{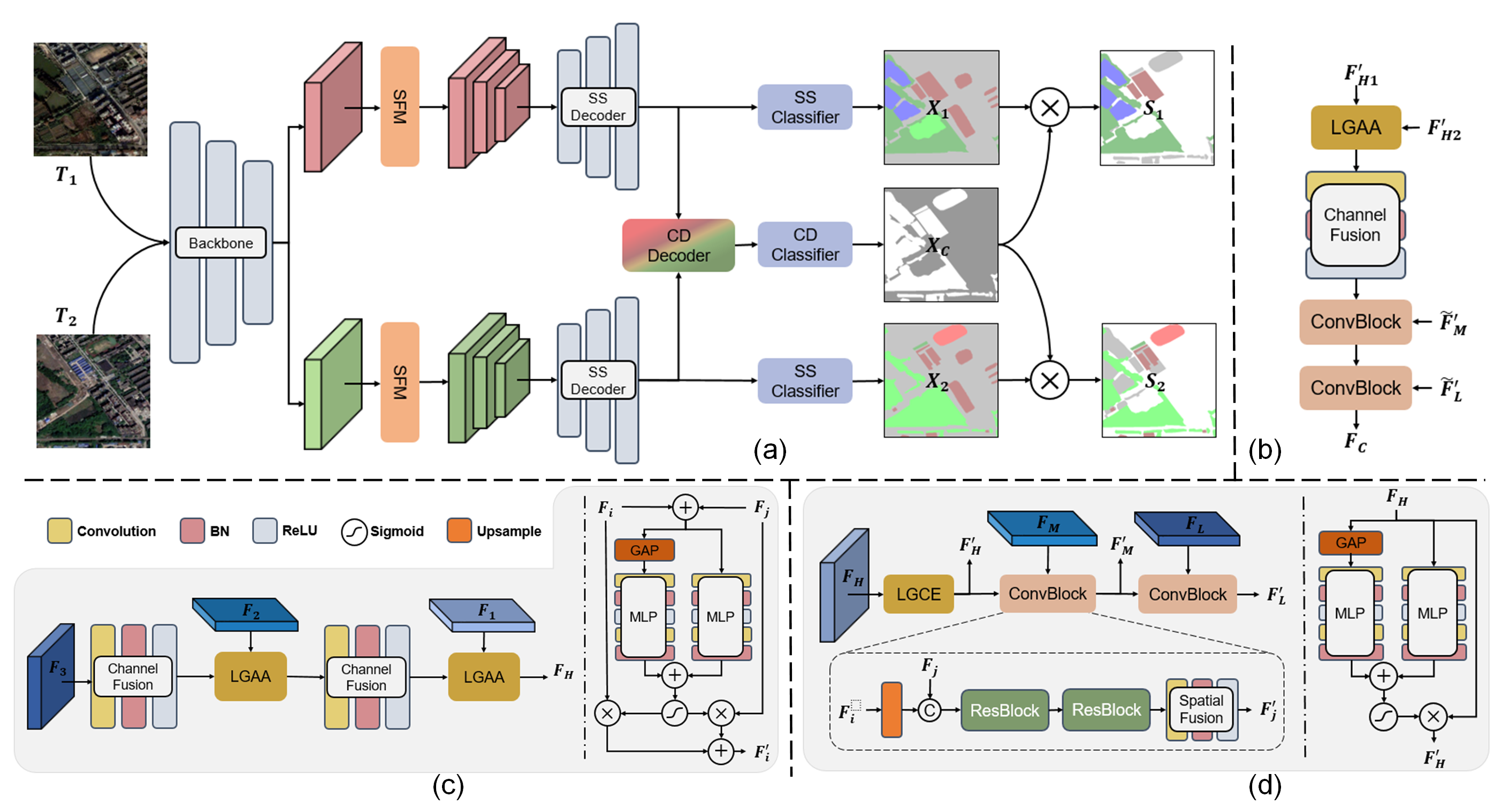}
        \caption{Architectures of our proposed LSAFNet and its components. (a) Flowchart of LSAFNet. (b) Architecture of CD Decoder. (c) Architecture of SFM and detailed structure of LGAA. (d) Architecture of SS Decoder and detailed structure of LGCE, respectively. 
        }
        \label{fig:LSAFNet}
\end{figure*}

According to where these triple branches communicate with each other, the aforementioned methods can be classified into two types, i.e. early-stage fusion models and middle-stage fusion models \cite{cheng2023changecomprehensivereview}. Early-stage fusion models like HRSCD \cite{daudt2019multitaskHRSCD} implement three encoders to extract features for each branch. The feature map inside BCD branch is obtained from scratch, hence fails to make good use of the semantics within dual SS branches. To better utilize the semantics, middle-stage fusion networks like \cite{BiSRNet, ding2024joint, DecoderFocus} capture change context on dual SS encoded features without specific BCD encoder. To construct dependencies between bitemporal images, these methods start correlating their triple branches from encoders. They either establish parallel BCD branch after SS encoders or jointly model semantic tokens for triple branches.

Though promising results have been achieved, we argue that their entangled design of triple branches may not be necessary and may not be optimal. For one thing, BCD can be interpreted as the ``exclusive or'' result of bitemporal semantic maps, thus it's possible to capture change context directly from late-stage bitemporal semantics while achieving satisfying accuracy. From another perspective, entangled design makes it harder to adapt pretrained foundation models into downstream SCD task to transfer their modeling capability in a plug-and-play paradigm due to intermediate feature entanglement \cite{li2024FoundationModelRSCD}. In this way, we propose a novel late-stage bitemporal feature fusion network with a shallow decoder interpreting change regions from SS decoded features. Fig. \ref{fig:architecture} shows the main difference between our proposed model and previous triple branches methods. Furthermore, \cite{daudt2018fully, BiSRNet, DecoderFocus} only apply naive fusion strategies like difference and concatenation when capturing change information without explicit change feature refinement. We argue that this is not sufficient for accurate change region localization and is vulnerable to irrelevant change semantics. To this end, we re-weight the primeval change features based on local and global context to boost representative ability.

The contribution of our work can be summarized as follows:
\begin{itemize}
\item A novel SCD method LSAFNet is proposed with more decoupled architecture of two branches of SS and one BCD branch. With dual SS branches only interact in late-stage, our network achieve satisfying accuracy while being friendly to foundation model implantation.

\item We propose LGAA module and LGCE module to refine features based on local and global context for better representative ability.

\item Comprehensive experiments are conducted on two public datasets, quantitative and qualitative studies show that our proposed LSAFNet outperforms state-of-the-art methods.
\end{itemize}

\section{Methods}
\label{sec:method}

\subsection{Overall Architecture}
\label{ssec:overview}
As depicted in Fig. \ref{fig:LSAFNet}(a), the whole architecture of our proposed LSAFNet follows a multitask learning paradigm, where the SCD is decoupled into two SS branches and a BCD branch. Given two input remote sensing images \(\mathit{T}_{1}\) and \(\mathit{T}_{2}\) carrying different temporal information, in the early stage of LSAFNet, we first model the intra-temporal LULC semantics through encoder-decoder architecture separately without any cross-temporal interaction. By applying a visual backbone network, we extract a series of feature maps denoted as \(\mathit{F}_{L}\), \(\mathit{F}_{M}\), \(\mathit{F}_{1}\), \(\mathit{F}_{2}\), and \(\mathit{F}_{3}\), with channel dimension of 64, 64, 128, 256, and 512, respectively. Then, \(\mathit{F}_{H}\) with more pivotal semantic information is obtained from \(\mathit{F}_{1}\), \(\mathit{F}_{2}\) and \(\mathit{F}_{3}\) through semantic fusion module(SFM). \(\mathit{F}_{L}\), \(\mathit{F}_{M}\) and \(\mathit{F}_{H}\) are further up-sampled and decoded layer by layer in the following SS decoder, and the corresponding semantic mask is predicted through its classifier. By now the semantic features from both temporal phases haven't meet each other, until the intermediate feature maps from SS decoders are aggregated inside BCD decoder. We utilize local-global attentional aggregation module to highlight the semantic differences across time interval while suppressing the irrelevant changes, and adopt cascaded convolution blocks to connect varied stages. Ultimately, a change region binary mask is obtained through change detection classifier, and we take it as guidance to mask out unchanged areas in both semantic masks to achieve the final semantic change predictions.

\subsection{Semantic Fusion Module}
\label{ssec:SFM}
The key to achieve satisfying SCD result lies in identifying and matching every pixel's semantic category between the given two input images from different temporal phases. Due to the intrinsic nature of remote sensing images having rich background context and varied object scales, the raw features extracted by backbone network suffers from the perplexity of inter-class similarity and intra-class variability. The different imaging periods of multi-temporal image series further bring in interference factors such as irrelevant seasonal and illuminating changes {\cite{BiFa}}. Therefore, it's crucial to construct more representative features to facilitate downstream SS and BCD subtasks. To this end, motivated by {\cite{DecoderFocus}}, we propose semantic fusion module(SFM) to aggregate features \(\mathit{F}_{1}\), \(\mathit{F}_{2}\) and \(\mathit{F}_{3}\) into a more representative feature map \(\mathit{F}_{H}\) layer by layer. After channel reduction through pointwise convolution, the lower-level feature map is aggregated with its next level counterpart in LGAA module. The process can be expressed as follows:
\begin{equation}
    \label{1}
    \mathit{F}_{2}^{'} = \rm{LGAA}(\mathit{F}_{2},{\rm{ReLU}}({\rm{BN}}({\rm{Con}}{{\rm{v}}_{1 \times 1}}(\mathit{F}_{3}))))
\end{equation}
\begin{equation}
    \label{2}
    \mathit{F}_{H} = \rm{LGAA}(\mathit{F}_{1},{\rm{ReLU}}({\rm{BN}}({\rm{Con}}{{\rm{v}}_{1 \times 1}}(\mathit{F}_{2}^{'}))))
\end{equation}
where BN represents the BatchNorm operation.

The proposed SFM distinguish from previous methods mainly on local-global attentional aggregation module(LGAA). Inspired by \cite{dai2021attentional}, we fuse adjacent levels of features with explicit per-channel re-weighting. By GAP, a global representative vector is obtained from the summation of two input feature maps. We further utilize two layers of Conv-BN-ReLU as the local channel
context aggregator, and combine the local and global channel context through addition. The re-weighting vectors are subsequently applied to their corresponding feature maps and the results are added as the output of LGAA. Take \(\mathit{F}_{i}\) and \(\mathit{F}_{j}\) as the LGAA's inputs, the output \(\mathit{F}_{i}^{'}\) can be expressed as follows:
\begin{equation}
    \label{3}
    {w_1} = {\rm{Conv(BN(ReLU(Conv(BN(GAP(}}{F_i} + {F_j}))))))
\end{equation}
\begin{equation}
    \label{4}
    {w_2} = {\rm{Conv(BN(ReLU(Conv(BN(}}{F_i} + {F_j})))))
\end{equation}
\begin{equation}
    \label{5}
    F_{i}^{'} = = {F_i}{\rm{Sigmoid}}({w_1} + {w_2}) + {F_j}(1 - {\rm{Sigmoid}}({w_1} + {w_2}))
\end{equation}
where GAP represents the global average pooling operation.

\subsection{Semantic Segmentation Decoder}
\label{ssec:SS Decoder}
As shown in Fig. \ref{fig:LSAFNet}(d), we use two parallel weight-sharing decoders for SS branches. The SS Decoder mainly comprises a local-global context enhancement module(LGCE) and two convolution blocks, gradually upsamples and aggregates adjacent levels of input feature maps for the final semantic map prediction and cross-temporal interaction.

Following the practice in {\cite{DecoderFocus}}, the ConvBlock consists of upsampling, concatenation and two cascaded ResBlocks to combine the two input features, and apply depthwise convolution to further fuse the spatial context. The high level feature containing pivotal semantics, denoted as \(\mathit{F}_{H}\), is first processed in LGCE to distinguish interested semantics from interference factors through channel attention. Similar to the calculation of LGAA, the output \(\mathit{F}_{H}^{'}\) of LGCE can be formulated as follows:
\begin{equation}
    \label{6}
    {w_1} = {\rm{Conv(BN(ReLU(Conv(BN(GAP(}}{F_H}))))))
\end{equation}
\begin{equation}
    \label{7}
    {w_2} = {\rm{Conv(BN(ReLU(Conv(BN(}}{F_H})))))
\end{equation}
\begin{equation}
    \label{8}
    F_{H}^{'} = = {F_H}{\rm{Sigmoid}}({w_1} + {w_2})
\end{equation}

\subsection{Change Detection Decoder}
\label{ssec:CD Decoder}
The above SS encoder-decoder branch only captures intra-temporal LULC categories within each temporal phase. To identify the changed region across two temporal phases and project intra-temporal LULC categories into cross-temporal change region semantics, we propose a simple yet efficient bridging decoder between two temporal branches in the late-stage of our network to achieve bitemporal interaction. The proposed CD Decoder, as depicted in  Fig. \ref{fig:LSAFNet}(b), receives three levels of decoded features from both SS Decoders and generate feature map \(\mathit{F}_{C}\) related to change regions. We first implement the same LGAA module in Sec. \ref{ssec:SFM} to merge high-level feature maps from both temporal branches. Then, we apply pointwise convolution to reduce its channel dimension and aggregates its semantic information with the spatial information from two subsequent lower-level feature maps layer by layer. For simplicity, we subtract one SS Decoder's output feature map from its counterpart of another SS Decoder and keep the absolute value as the lower-level features being processed in ConvBlocks. Given \(\mathit{F}_{H1}^{'}\), \(\mathit{F}_{M1}^{'}\), \(\mathit{F}_{L1}^{'}\) from \(\mathit{T}_{1}\) SS Decoder, and \(\mathit{F}_{H2}^{'}\), \(\mathit{F}_{M2}^{'}\), \(\mathit{F}_{L2}^{'}\) from \(\mathit{T}_{2}\) SS Decoder, the \(\mathit{F}_{C}\) can be calculated as follows:
\begin{equation}
    \label{9}
    {\widetilde {F}_{H}} = {\rm{LGAA(}}F_{H1}^{'} + F_{H2}^{'})
\end{equation}
\begin{equation}
    \label{10}
    \widetilde F_{H}^{'} = {\rm{ReLU(BN(Con}}{{\rm{v}}_{1 \times 1}}({\widetilde F_{H}})))
\end{equation}
\begin{equation}
    \label{11}
    \widetilde F_{M}^{'} = \left| {F_{M1}^{'} - F_{M2}^{'}} \right|
\end{equation}
\begin{equation}
    \label{12}
    \widetilde F_{L}^{'} = \left| {F_{L1}^{'} - F_{L2}^{'}} \right|
\end{equation}
\begin{equation}
    \label{13}
    {F_{C}^{'}} = f(\widetilde F_H^{'},\widetilde F_M^{'})
\end{equation}
\begin{equation}
    \label{14}
    {F_{C}} = f(F_{C}^{'}, \widetilde F_L^{'})
\end{equation}
where \(\mathit{f(\cdot)}\) represents ConvBlock described in Sec. \ref{ssec:SS Decoder}, \(\mathit{\left|  \cdot  \right|}\) represents the absolute value operator.

\subsection{Loss Function}
\label{ssec:loss}
Our multitask schemed network produces three prediction maps in total, namely \(\mathit{X}_{1}\), \(\mathit{X}_{2}\), and \(\mathit{X}_{C}\). The \(\mathit{X}_{1}\) and \(\mathit{X}_{2}\) serves as the LULC semantic maps correspond to each temporal phase, while \(\mathit{X}_{C}\) is a binary change mask, denoting the changed regions across time. 
In this work we supervise over \(\mathit{X}_{1}\), \(\mathit{X}_{2}\) and \(\mathit{X}_{C}\) instead of semantic change maps. We choose the multi-class cross-entropy loss for semantic maps optimization and the binary cross-entropy loss for change region supervision. The formulation of \(\mathcal{L_{SS}}\) and \(\mathcal{L_{BCD}}\) can be expressed as
\begin{equation}
    \label{15}
    \mathcal{L_{SS}} = - \frac{1}{N}\sum\limits_{i = 1}^N {{y_i}{\rm{log}}({p_i})}
\end{equation}
\begin{equation}
    \label{16}
    \mathcal{L_{BCD}} = - {y_c}{\rm{log}}({p_c}) - (1 - {y_c})\log (1 - {p_c})
\end{equation}
where \(\mathit{N}\) represents the number of categories in the semantic maps, \(\mathit{y}_{i}\) and \(\mathit{p}_{i}\) represents the groundtruth label index and the predicted probability of each category respectively, and \(\mathit{y}_{c}\) and \(\mathit{p}_{c}\) represents the groundtruth label index and the corresponding predicted probability of change region in the binary change map. We ignore the no-change class in the semantic change labels to maintain the semantic category consistency between semantic change labels and predicted masks. To better align the bitemporal SS subtask and BCD subtask, a semantic consistency loss \(\mathcal{L_{SC}}\) is proposed in {\cite{BiSRNet}} as 
\begin{equation}
    \label{17}
    \mathcal{L_{SC}} = \left\{ {\begin{array}{*{20}{c}}
{1 - \cos (x_1,x_2),{y_c} = 1}\\
{\cos (x_1,x_2),{y_c} = 0}
\end{array}} \right.
\end{equation}
where \(\mathit{x}_{1}\) and \(\mathit{x}_{2}\) signify the feature vectors of a pixel in \(\mathit{X}_{1}\) and \(\mathit{X}_{2}\) respectively. The total loss \(\mathcal{L}\) implemented through this paper is defined as follows:
\begin{equation}
    \label{18}
    \mathcal{L} = \mathcal{L_{BCD}} + 0.5\times (\mathcal{L_{SS\rm1}} + \mathcal{L_{SS\rm2}}) + \mathcal{L_{SC}}
\end{equation}

\section{Experiments} 
\label{sec:experiment}

\subsection{Datasets and Evaluation Metrics}
\label{ssec:data}
To verify the effectiveness of our model, we conduct experiments on two publicly available SCD dataset, including SECOND {\cite{SECOND}} and Landsat-SCD {\cite{Landsat}}. SECOND dataset consists of 2968 pairs of bitemporal images of size 512 $\times$ 512, with resolution ranging from 0.5m to 3.0m, 
including building, water, tree, low vegetation, ground and playground. 
Landsat-SCD dataset comprises 8468 pairs of bitemporal images of size 416 $\times$ 416, with a consistent resolution of 30m, 
including water, farmland, building and desert.

For fair comparison, we keep the same scaling and partition strategy as previous work \cite{BiSRNet, ding2024joint, DecoderFocus} throughout the whole experiment. To quantitatively measure the similarity between the predicted bitemporal semantic change probability maps and their corresponding labels, we introduce four well-established indicators, including mIoU and Avg evaluate the overall segmentation performance, as well as SeK and \(\mathit{F_{scd}}\) specifically focus on the semantic discrimination within changed regions.

\subsection{Implementation Details}
\label{ssec:implementation}
Experiments are conducted with PyTorch on two NVIDIA RTX4090 GPUs. We deploy ResNet-34 as backbone, initialize our network with Kaiming Initialization \cite{he2015delvingKaimingInitialization} and train it for 50 epochs with batch size of 8. SGD with weight decay of 5e-4 and momentum factor of 0.9 is selected as the optimizer, with initial learning rate set to 0.07. Common data augmentation including random flipping and rotation is carried out during training. Code are publicly available at \href{https://github.com/STORMTROOPERRR/RSISCD}{https://github.com/STORMTROOPERRR/RSISCD}.

\begin{figure}[htbp]
    \vspace{-0.5em}
        \centering
        \includegraphics[width=0.48\textwidth]{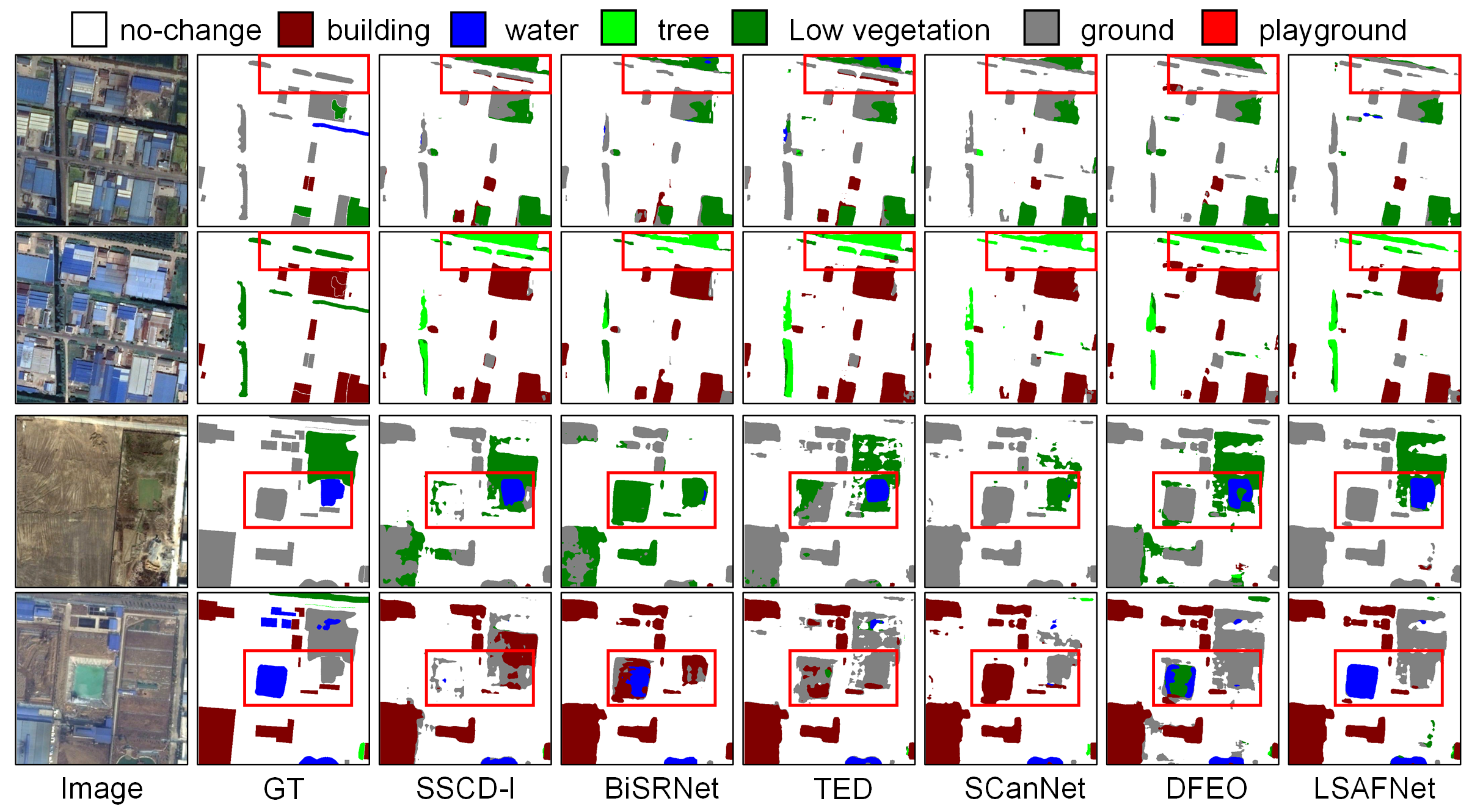}
        \caption{Qualitative comparisons of the results on SECOND dataset. First two rows and last two rows contain different bitemporal image pairs, respectively.}
        \label{fig:SECOND}
\end{figure}

\subsection{Comparison and Analysis}
\label{ssec: comparison}
\begin{figure}[htbp]
    \vspace{-0.5em}
        \centering
        \includegraphics[width=0.48\textwidth]{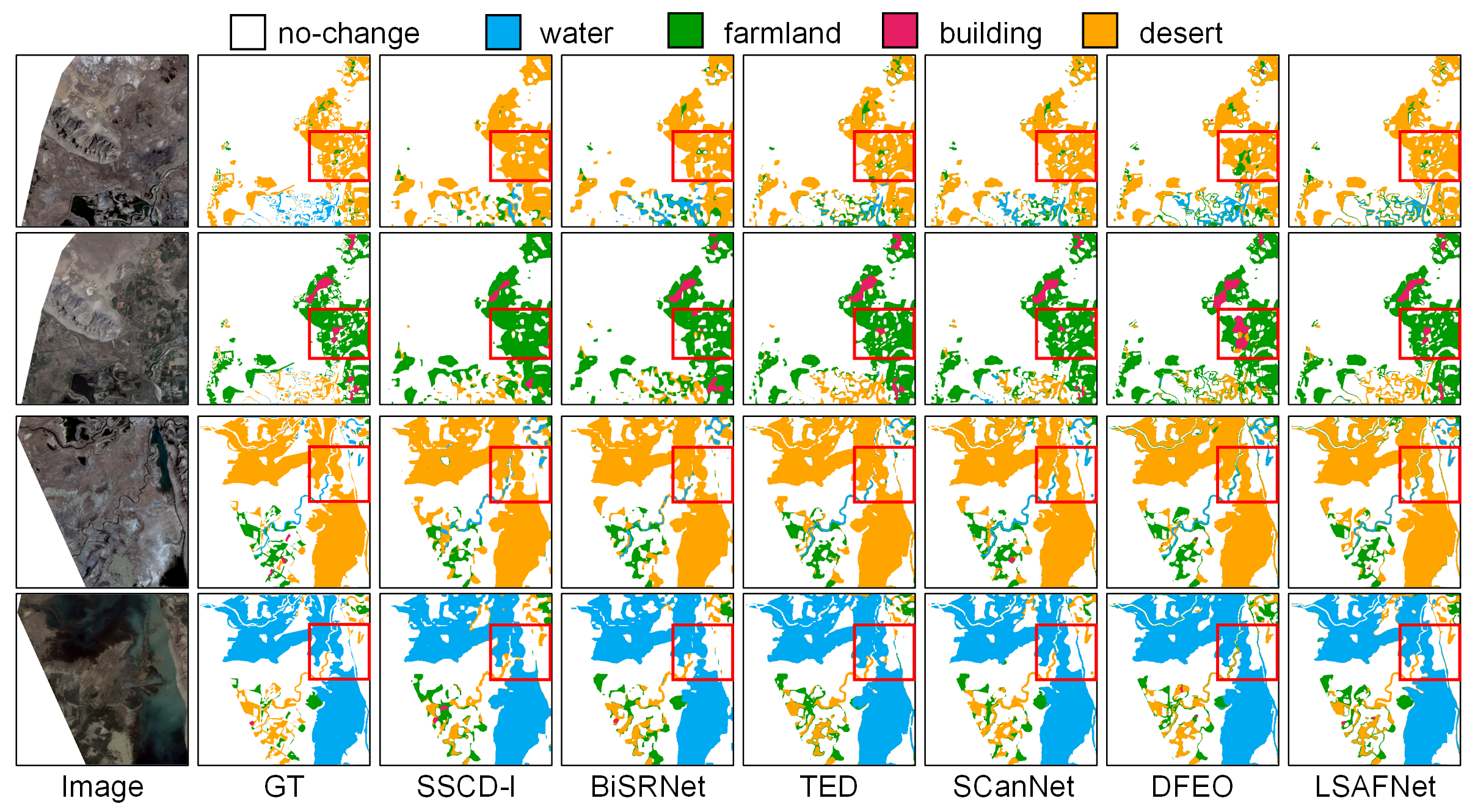}
        \caption{Qualitative comparisons of the results on Landsat dataset. First two rows and last two rows contain different bitemporal image pairs, respectively.}
        \label{fig:Landsat}
\end{figure}

We compare our proposed model with other state-of-the-art methods for performance evaluation. Quantitative results are listed in Table \ref{tab:table1} and Table \ref{tab:table2} for each dataset, with best results marked in \textbf{bold}. Statistics show that our proposed model achieve new SOTA results on both dataset, especially on changed regions across time. For more intuitive demonstration, we select two pairs of bitemporal images from both datasets to qualitatively evaluate different models' performance in Fig. \ref{fig:SECOND} and Fig. \ref{fig:Landsat}. Note that for simplicity, we only display the best six of all competing methods. We further highlight key areas where different models perform most diversely with red box. Fig. \ref{fig:SECOND} shows that our model can achieve high intra-category consistency with strong semantic capture ability. Fig. \ref{fig:Landsat} reveals our model's promising capability of modeling changed regions with various scales and delicate contour. Though being more decoupled, our proposed BCD branch succeeds in capture the change context between bitemporal images, while two SS branches maintain high accuracy in modeling each temporal phase's LULC semantics. The local global context aggregation guides SS encoders to extract representative features, ensuring the satisfying performance for both SS and BCD subtasks.

\begin{table}[htbp]
  \centering
  \caption{Numerical Results of Different Models on SECOND}
    \begin{tabular}{ccccc}
    \toprule
    Method & mIoU(\%) & Avg(\%) & SeK(\%) & Fscd(\%) \\
    \midrule
    FC-Siam-conv & 68.86  & 86.92  & 16.61  & 56.45  \\
    FC-Siam-diff & 68.96  & 86.86  & 16.50  & 56.23  \\
    HRSCD & 71.15  & 86.62  & 18.80  & 58.21  \\
    SCDNet & 70.95  & 87.29  & 19.75  & 59.77  \\
    SSCD-l & 72.60  & 87.19  & 21.86  & 61.22  \\
    Bi-SRNet & 73.38  & 87.48  & 22.43  & 61.62  \\
    TED   & 73.05  & 87.20  & 22.37  & 61.23  \\
    SCanNet & 73.20  & 87.46  & 23.34  & 62.59  \\
    DEFO-MLTSCD & 73.76  & \textbf{87.80} & 23.73  & 62.73  \\
    LSAFNet (Ours) & \textbf{74.01} & 87.66  & \textbf{24.32} & \textbf{63.20} \\
    \bottomrule
    \end{tabular}%
  \label{tab:table1}%
\end{table}%

\subsection{Ablation Study}
\label{ssec:ablation}

To quantitatively measure the improvements brought by each core component of our proposed model, we further conduct a series of ablation study on two datasets. We define our base model as the proposed LSAFNet without LGAA module and LGCE module. To deactivate LGAA module, we first concatenate two input features and apply a MLP to perform channel reduction. We replace LGCE module with identity when needed. Experimental results in Table \ref{tab:ablation} suggest that our proposed LGAA module and LGCE module both have its own contribution to the overall performance of our proposed LSAFNet. LGAA module is utilized both in SS encoders and BCD decoder, thus having a major impact on elevating our proposed model's capability.

\begin{table}[htbp]
  \centering
  \caption{Numerical Results of Different Models on Landsat-SCD}
    \begin{tabular}{ccccc}
    \toprule
    Method & mIoU(\%) & Avg(\%) & SeK(\%) & Fscd(\%) \\
    \midrule
    FC-Siam-conv & 79.31  & 90.79  & 36.11  & 76.04  \\
    FC-Siam-diff & 77.68  & 88.53  & 32.75  & 73.89  \\
    HRSCD & 78.51  & 91.47  & 32.90  & 73.20  \\
    SCDNet & 80.14  & 93.62  & 40.05  & 75.17  \\
    SSCD-l & 79.33  & 92.36  & 41.43  & 75.84  \\
    Bi-SRNet & 82.19  & 93.16  & 40.09  & 76.01  \\
    TED   & 84.22  & 93.98  & 45.60  & 78.47  \\
    SCanNet & 85.19  & 94.07  & 49.33  & 80.52  \\
    DEFO-MLTSCD & 87.49  & 94.32  & 49.26  & 81.39  \\
    LSAFNet (Ours) & \textbf{87.60} & \textbf{94.46} & \textbf{49.94} & \textbf{81.66} \\
    \bottomrule
    \end{tabular}%
  \label{tab:table2}%
\end{table}%

\begin{table}[htbp]
  \vspace{-1.0em}
  \centering
  \caption{Ablation Study on Two Datasets}
    \begin{tabular}{ccccc}
    \toprule
    \multirow{2}[2]{*}{Method} & \multicolumn{2}{c}{SECOND} & \multicolumn{2}{c}{Landsat-SCD} \\
          & mIoU(\%) & SeK(\%) & mIoU(\%) & SeK(\%) \\
    \midrule
    Base  & 73.59 & 23.37 & 86.09 & 48.67 \\
    Base + LGAA & 73.82 & 23.90 & 87.45 & 49.57 \\
    Base + LGCE & 73.69 & 23.69 & 87.44 & 49.29 \\
    LSAFNet (Ours) & \textbf{74.01} & \textbf{24.32} & \textbf{87.60} & \textbf{49.94} \\
    \bottomrule
    \end{tabular}%
  \label{tab:ablation}%
  \vspace{-0.1em}
\end{table}%

\section{Conclusion}
\label{sec:conclusion}

In this letter, to design a multi-task learning based SCD network in a more disentangled manner, we propose a novel late-stage bitemporal feature fusion network LSAFNet that only bridge bitemporal features in decoder stage. To extract more representative features in dual SS encoders, we proposed LGAA module to refine feature maps through aggregated local and global context re-weighting, and further utilize it to highlight change context across time while suppressing irrelevant changes. We further propose LGCE module to enhance the high-level features in SS decoders to boost LULC semantics modeling. Experiments on two public datasets verify our model's effectiveness, and ablation study confirms each component's contribution. We will harness our proposed architecture's disentanglement strengths to adapt pretrained foundation models into SCD field in our future work.





\ifCLASSOPTIONcaptionsoff
  \newpage
\fi

{\small
\bibliographystyle{IEEEtran}
\bibliography{refs}
}

\end{document}